\documentclass[12pt, a4paper]{article} 
\usepackage[utf8]{inputenc}
\usepackage{dcolumn,lscape}
\usepackage{amsmath,longtable,multicol,dcolumn,tabularx,graphicx,amssymb}
\usepackage{exscale,amsthm,multirow,rotating}
\usepackage{bbm}
\usepackage{tikz,dsfont,float}
\usepackage[caption = false]{subfig}
\usetikzlibrary{calc}
\usetikzlibrary{positioning}

\usepackage{graphicx}
\usepackage{amsmath,amssymb,dsfont, mathtools}
\usepackage{makecell}
\usepackage{multirow}
\usepackage{exscale,longtable,multicol,dcolumn,tabularx,bm}
\usepackage{adjustbox}
\usepackage{booktabs} 
\usepackage{mathptmx}      
\usepackage{mathrsfs}
\DeclareMathOperator{\EX}{\mathbb{E}}
\usepackage{caption}
\usepackage{xcolor, ulem}
\begin{document}

\def\spacingset#1{\renewcommand{\baselinestretch}%
{#1}\small\normalsize} \spacingset{1}


  \title{\bf Statistical learning for change point and anomaly detection in graphs}
  \author{Anna Malinovskaya\\
  	\small{Leibniz University Hannover, Germany}\\
  	Philipp Otto\\
  	\small{University of G\"{o}ttingen, Germany}\\
  	Torben Peters\\
  	\small{Leibniz University Hannover, Germany}}
  
  \maketitle
\begin{abstract}
Complex systems which can be represented in the form of static and dynamic graphs arise in different fields, e.g. communication, engineering and industry. One of the interesting problems in analysing dynamic network structures is to monitor changes in their development. Statistical learning, which encompasses both methods based on artificial intelligence and traditional statistics, can be used to progress in this research area. However, the majority of approaches apply only one or the other framework. In this paper, we discuss the possibility of bringing together both disciplines in order to create enhanced network monitoring procedures focussing on the example of combining statistical process control and deep learning algorithms. Together with the presentation of change point and anomaly detection in network data, we propose to monitor the response times of ambulance services, applying jointly the control chart for quantile function values and a graph convolutional network.
\end{abstract}

\noindent%
{\it Keywords:} Network Monitoring, Statistical Process Control, Control Charts, Machine Learning on Graphs, Graph Convolutional Networks.
\vfill

\spacingset{1.45} 

\section{Introduction}
\label{intro}
Network representation is fascinating. It conveys complexity by introducing a relational structure between objects and enables incorporation of various information. The field was founded in 1735 when Leonhard Euler solved the K\"{o}nigsberg bridge problem, and since then network science has developed into a significant area of study. The broad interest of the statistical community in graph-based data analysis arose in the last century with the development of Erd\"{o}s-R\'{e}nyi-Gilbert model (\cite{erdoes1959graphs}, \cite{gilbert1959graphs}) introducing a probabilistic view on the problem. Another significant factor in the growing popularity of networks is the availability of extensive data sources. The present era of Big Data provides a unique opportunity to gain remarkable insight into molecular, social, economic and many other structures (cf. \cite{grennan2014molecular}, \cite{o2008analysis}, \cite{schweitzer2009economic}). Consequently, it shifts perspective to new analytical methods, which are not solely developed in the traditional statistical framework but also involve recent inventions in machine learning. Although the integration of artificial intelligence in network data analysis has achieved impressive results, the research in this area is in its early stages. Many existing machine learning algorithms cannot be applied directly to graphs because of their specific structural properties. 
To be precise, a standard vector representation of graphs does not exist compared to such data types as images or audio which can be defined on regular lattices. This and other aspects related to the methods' generalisation and their evaluation challenge the development of machine learning approaches for network data but also offer the possibility to identify and unify the benefits of artificial intelligence and the traditional statistical framework for analysing graph-structured data efficiently. 

Considering dynamic networks, those which develop over time, and one of the main interests in their study -- the detection of anomalous behaviour many powerful techniques based on statistical inference exist to perform network monitoring. However, the identification of the subsequent timestamp when the network system started deviating from its target state is not a solution to the issue as the inspection and possible improvement of the system are the following necessary steps to undertake. 
If we decide to perform monitoring completely by a machine learning algorithm which could perform the inspection step, we might encounter certain obstacles. Possible problems include the detection speed, data amount, interpretability and reliability, which remain poorly understood. Probably, it will also overcomplicate the monitoring process, especially in the case when the network is stable and does not experience any sudden changes. However, a joint approach could combine benefits from both classical and modern learning foundations, and as a result improve network surveillance and statistical learning in general. The expression "statistical learning" is a broader term for defining the approaches to learn from data, including algorithms based on the artificial intelligence as well as methods which are purely statistical (cf. \cite{hastie2009elements}). 

An example of a well-known monitoring tool from the classical statistical framework is the control chart. It belongs to Statistical Process Control (SPC) being an effective instrument for detecting process deviation from the in-control or target state. Its universality and efficient technique to monitor the process on-line cannot be easily outperformed by novel approaches. However, as soon as a chart detects a change, often the investigation of a possible reason happens manually. If we consider the surveillance of graph-structured data that can be considerably voluminous and challenging to process in its raw format, the task to identify the cause of the signal and, if applicable, to resolve the issue may become extremely time-consuming. To improve the actions in the post-monitoring phase, we propose an enhanced application of the control charts which, in case of a signal, is followed by a graph machine-learning algorithm that can operate on graphs to classify the cases, i.e., identify the reasons which led to the out-of-control state.

It is worth mentioning that the idea to extend the functionality of the control chart by combining it with machine learning algorithms is already an existing approach. There are several publications which propose different possibilities to bring together the two areas and generally evaluate the usage of machine learning in SPC (cf. \cite{kang2011spcml}, \cite{psarakis2011spcml}, \cite{psarakis2011bspcml}, \cite{diren2019spcml}, \cite{khoza2019spcml}, \cite{zan2019spcml}, \cite{apsemidis2020spcml}). However, the authors are not aware of the introduction to the topic from the point of network data monitoring which currently obtains rapidly growing attention.    

In this paper, we propose a monitoring method which consolidates the control chart for quantile function values and a graph convolutional network. In Section \ref{sec:2} we introduce the mathematical definition of a graph, following by the description of the change point and anomaly detection problems with the respective literature overview in Section \ref{sec:3}. Section \ref{sec:4} focuses on the advancements of the graph learning representation including the description of graph convolutional networks. The simulation study of monitoring compliance with the response time prescribed to ambulance services is described in section \ref{sec:5}.  We conclude with a discussion of possibilities to expand the joint applications of machine learning with the classical statistical tools and present several directions for future research.

\section{What is a graph?} 
\label{sec:2}
A graph (interchangeably called ``network'') $G = (V, E)$ is defined by nodes (also known as ``vertices'') $v_{i} \in V$ and edges $ e_{i,j} \in E$ with $e_{i,j}$ being an edge (also called ``link'' or ``tie'') between vertices $v_i$ and $v_j$. Usually, the network is represented by a binary or weighted adjacency matrix $\bm{A} \in \mathbb{R}^{|V|\times |V|}$, where $|V|$ represents the total number of nodes. Two vertices are adjacent if they are connected by an edge. If we consider a binary adjacency matrix, then $A_{ij} = 1$, otherwise, $A_{ij} = 0$. In case of an undirected network, $\bm{A}$ is symmetric. Additionally, we can assign nodal or edge attributes which are described by $\bm{X}^V$ and $\bm{L}^E$ so that $\bm{x}_i$ defines attributes of node $v_i$ and $\bm{l}_{i,j}$ contains attributes of edge $e_{i,j}$. If the graph is weighted, we can incorporate the weights as one of the edge attributes into $\bm{L}^E$ or into the representation of $\bm{A}$ directly. For the purpose to distinguish between a network as graph-structured data and the neural network approach, we use a full name for the latter, e.g. ``graph convolutional network'' or ``neural network''.

Since graphs are powerful abstractions, there are numerous applications of them, including semantic, transportation, document citation, protein-protein interactions networks and many others (cf. \cite{chen2015semanticnet}, \cite{brevier2007proteinnet}). Consequently, the focus of the statistical analysis of networks differs from concentrating on the descriptive properties of the graph up to implementing inferential modelling and beyond.  
%

\section{Change point and anomaly detection in network data}
\label{sec:3}
In this section, we demonstrate the application of network modelling and other approaches which enable to analyse the graph-structured data over time. To be precise, we discuss the change point and anomaly detection in network data. 
\subsection{What is a change point?}

Network monitoring is a form of online surveillance procedure to detect a change point when the network system starts deviating from a so-called in-control state, i.e., the state when no unaccountable variation of the process is present. In other words, consider $\bm{s}_t = (s_1(G_t), \dots, s_p(G_t))'$ as a collection of $p$ network statistics which are derived from $G$ at timepoint $t$. Following, let $F_{0}$ be the in-control or target distribution and $F_{\tau}$ the out-of-control distribution.\newpage We call $\tau$ a changepoint for a stochastic process $\bm{s}_t$, if 
\begin{equation*}\label{eq:cp_model}
\bm{s}_t \quad \sim \quad \left\{ \begin{array}{cc}
F_0 &  \text{ if } t < \tau \\
F_{\tau} & \text{ if } t \geq \tau \\
\end{array} \right. \, .
\end{equation*}

\subsection{Methods for network monitoring}
The approaches to monitor network data can be mainly subdivided into hypothesis testing methods, Bayesian methods and scan methods. The first category is dominated by the application of different forms of control charts which represent the leading SPC method. Typically, a control chart consists of the control statistic depending on one or more characteristic quantities, plotted in time order, and one or two horizontal lines known as control limits. There is a strong connection between control charts and hypothesis testing, as it repeatedly tests at different points of time $t$ the null
hypothesis $H_{0, t}$ against the alternative $H_{1, t}$. If we are interested in monitoring the deviation of the network statistics $\bm{s}_t$ from its expected value $\bm{s}_0$, then we specify
\begin{equation*}
H_{0, t}: \, \EX(\bm{s}_t) = \bm{s}_{0} \qquad \text{against} \qquad H_{1, t}: \, \EX(\bm{s}_t) \neq \bm{s}_{0} \, .
\end{equation*}
A hypothesis $H_{0, t}$ is rejected if the control statistic is equal to or exceeds the value
of the control limit.
Usually, application of control charts is divided into Phase I and Phase II, which have two distinct objectives. The data collected in Phase I serves as a baseline for estimation of the parameters such as expected value and calculation of control chart limits. In other words, we calibrate the control chart based on the observations which were collected under the assumption that the process is in its target state. In Phase II we start monitoring the system which is assumed to stay in-control and examine the functionality of the control chart in respect to detected anomalies, i.e., out-of-control states, and to false alarms -- when no abnormality is presented but the chart signals a change.

There are several ways to classify control charts. In terms of the number of variables, there are univariate ($p = 1$) and multivariate ($p>1$) control charts. For both types there exist various examples of applications in network surveillance. For instance, McCulloh and Carley \cite{mcculloh2011militnets} monitor the topology statistics of military networks applying the Cumulative Sum (CUSUM) chart. The Shewhart and Exponentially Weighted Moving Average (EWMA) charts were used by Wilson et al. \cite{wilson2019dcsbm} in combination with the dynamic Degree-Corrected Stochastic Block Model (DCSBM) to generate the networks and then perform surveillance over the Maximum Likelihood (ML) estimates. The application of EWMA and
CUSUM to degree measures for detecting outbreaks on a weighted undirected network was introduced in \cite{hosseini2018charts}. Sparks and Wilson \cite{sparks2018ewma} detect communication outbreaks designing an adaptive EWMA control chart. Malinovskaya and Otto \cite{mal2020tergm} apply multivariate EWMA (MEWMA) and CUSUM (MCUSUM) together with the Temporal Exponential Random Graph Model (TERGM) and detect the beginning of the national lock-down period due to the COVID-19 pandemic by monitoring daily flights in the United States. Farahani et al. \cite{farahani2017overview} evaluate the combination of the former charts used with the Poisson regression model for monitoring social networks. An overview of further control chart-based studies can be found in \cite{noorossana2018overview}. Regarding other hypothesis testing methods, Azarnoush et al. \cite{azarnoush2016lrt} propose monitoring network attributes instead of measures derived from its connectivity structure applying a logistic regression model and a likelihood-ratio test. Another method that shares similar assumptions as \cite{azarnoush2016lrt} and incorporates vertex attributes is described in \cite{miller2013efficient}.

The Bayesian framework presented by Heard et al. \cite{heard2010bayes} applies a two-stage approach using control chart limits based on a Bayesian predictive distribution. This
type of methods concentrates on identifying anomalous behaviour between pairs of
nodes (stage 1) which later are monitored as a sub-network (stage 2). Technically, for each
pair of vertices, a communication trend is developed with the increments of the
process following a Bayesian probability model. The node pair is considered as anomalous if the $\mathit{p}$-value, which is derived throughout time, falls below the defined threshold (e.g. 0.05). Scan-based monitoring is known in the engineering literature as ``moving window analysis'' and is based on the concept of scanning a particular region of data by calculating a standardised metric for each window. This idea is applied in $\cite{priebe2005scan}$ for detecting anomalies in the directed graphs (digraphs). Woodall et al. \cite{woodall2017graphs} provide a broader discussion of these both categories.

\subsection{How can we specify anomaly detection in terms of network monitoring?}
It is often the case that a monitoring statistic is aggregated from several observations which were collected within a specific time frame. That means, if the change point is detected, it is possible that only a few samples were anomalous and the rest not. Thus, we would need to perform anomaly detection as a postprocessing step to resolve a possible issue in the network system.

It is worth mentioning that there is no unique definition of the problem ``anomaly detection''. Akoglu et al. \cite{akoglu2015elements} uses ``change point detection'' as a synonym for the anomaly detection problem for dynamic graphs, emphasising the existing difference between methods for static and temporal graphs in the survey. On the contrary, Ranshous et al. \cite{ranshous2015overview}, who also provide an extensive methodological overview, introduce the change point detection as a subcategory of anomaly detection problem. The reason of considerably different point of views is that the meaningful definition can only be established after a context and particular application are specified, otherwise, it is ambiguous. Here, under ``anomaly'' we understand an abnormal activity being a sudden and a significant change in the interaction patterns of a network \cite{zhao2018anomaly}. Consequently,  ``anomaly detection'' defines the task to find the networks which significantly differ from the majority of the reference networks and, if applicable, to ascertain the type of anomalous behaviour.

In contrast to the introduced definition of an anomalous observation in form of a whole graph, we can also define the anomaly detection problem in terms of edges or vertices, i.e., the aim is to find a subset of nodes/edges such that every element in this subset presents an uncommon evolution compared to other nodes/edges in the network. Another possible task is to identify anomalous subgraphs.  Recent advancements in the area of machine learning on graphs led to impressive results in solving the specified problems by applying the graph convolutional network (GCN) framework (cf. \cite{wang2020ocgnn}, \cite{zheng2019addgraph}, \cite{kumagai2020gcngraph}). However, to extract necessary information and provide substantial results, the neural network needs the graph data to be constructed as a set of features (called ``embeddings'') without neglecting the relational structure and corresponding attributes. This task can be fulfilled by the graph representation learning techniques, which are briefly discussed together with the GCN in the following section.
%

\section{Graph representation learning} 
\label{sec:4}
Undeniably, the hand-engineered graph statistics $\bm{s}_t$ are useful in analysing the graph-structured data in terms of interpretation and computational costs. However, the manual selection of which features should be incorporated into the metrics, and further determination of statistics can be a time-consuming process. Moreover, this approach is restrictive because neither the selection of features nor of metrics can be adapted through a learning process, which crucially constrains the effectiveness of machine learning-based algorithms. An alternative to encode network structure compactly and without losing any relevant information is Graph Representation Learning (GRL). It is worth mentioning that in contrast to conventional methods, where we see the selection and design of graph statistics as a preprocessing step, the GRL techniques regard the problem to learn embeddings as a machine learning task. To be more precise, the goal is to learn and optimise a mapping that embeds vertices, edges or entire (sub)graphs as points in a low-dimensional vector space $ \mathbb{R}^d$ such that geometric relationships in this latent space reflect the structure of the initial graph \cite{hamilton2017represgraph}. Subsequently, the learned representation can be used as input for the main machine learning task such as classification.  Hamilton \cite{hamilton2020represgraph} comprehensively reviews traditional as well as modern learning approaches over graph-structured data.

If we consider node embedding, the main purpose is to find a projection
\begin{equation*}
f_{\Theta}: v_i \rightarrow \bm{z}_i \in \mathbb{R}^d,
\end{equation*}
where $d \ll N$ and $\bm{z}_i = \{z_1, z_2, \dots, z_d\}$ represents the embedded vector that captures the node $v_i$ graph position and the structure of its local graph neighborhood, and $f_{\Theta}$ is a mapping function parametrised by $\Theta$. Depending on the embedding method, the incorporation of edge and nodal attributes into the latent representation $\bm{z}_i$ of the node $v_i$ is also possible. Further encoding techniques which do not only focuss on node representation together with the discussion of recent challenges in GRL can be found in \cite{hamilton2017represgraph}, \cite{cai2018represgraph}, \cite{chen2020represgraph} and \cite{gogoglou2020represgraph}.

\subsection{Shallow node embedding methods}

In general, the node embedding techniques can be subdivided into shallow and deep learning-based methods. The shallow embedding approaches define the mapping function $f$ (also known as ``encoder'') as simply being an embedding-lookup. In this case, the set of trainable parameters is optimised directly, meaning that $\Theta = \bm{Z}$, with $\bm{Z} \in \mathbb{R}^{d\times |V|}$ being a matrix, where each column defines node embeddings for vertices $V$. The best-known techniques are either based on matrix factorization (e.g. Laplacian eigenmaps) or random-walk statistics (e.g. DeepWalk and node2vec) \cite{hamilton2017represgraph}.
However, shallow embedding approaches have some considerable limitations. The first issue is that there is a unique embedding for each node in the graph, meaning that no parameters are shared across vertices resulting in the absence of regularisation. Another problem is the ability to generate embeddings only for nodes that were present during the learning process. That means, the graph structure should remain unchanged for the method to work correctly, which is highly unrealistic in many applications. Also, the embeddings reflect only the structural information, i.e., the incorporation of the node or edge attributes is not possible.

\subsection{Node embedding methods based on deep learning}
To overcome the discussed limitations, another framework based on deep learning techniques was proposed. Before starting with its introduction, it is important to define what ``deep learning'' means. ``Deep learning'' is a group of machine learning algorithms which is able to learn gradually a large number of parameters in an architecture composed by multiple non-linear transformations, e.g. neural networks with a multilayer structure. A general overview of methods including the principles of constructing and training a neural network model can be found in \cite{calin2020deeplearning}. In terms of graphs, the adaptation of such algorithms would enable us to generate a representation of nodes that would depend on both the structure of the graph and additional feature information. The pioneer framework is known as ``Graph Neural Networks'' (GNNs) \cite{scarselli2008graph} which establishes the idea of including the neighbourhood information of a node into its latent representation, applying neural message passing form. The earliest GNN variations were limited in covering edge features and also were restricted in the choice of trainable parameters. Consequently, many advanced forms of GNNs arose, the best-known example being Graph Convolutional Networks (GCNs) \cite{kipf2017graph}. Both GNNs and GCNs belong to a more general category which is Message Passing Neural Networks (MPNNs) \cite{gilmer2017graph}. For readers who are interested in further neural network architectures which operate on graphs, we suggest \cite{wu2020comprehensive} as a reference. 

\subsubsection{Two application phases of Graph Convolutional Networks}


The application of GCNs is structured into two main phases: message passing phase and readout phase. The goal of the first phase is to propagate the information across the nodes in order to create a new representation of the whole graph. In the readout phase, the obtained graph representation is used to solve a particular task. As we can see, there are direct similarities with the description of two steps defined for any GRL techniques at the beginning of Section \ref{sec:4}.

In the first phase, consider $k = 0, \dots, K$ to be the number of message passing iterations. Later we will see that $K$ equals the number of graph convolutional layers in the neural network. Next, we define $N(i)$ to contain the neighbourhood nodes of $v_i$ and $\bm{h}^k_{i}$ to be a hidden embedding of node $v_i$ incorporating the aggregated information (called ``message'') from $N(i)$ at the interation $k$. Initially at $k = 0$,   $\bm{h}^0_{i} = \bm{x}_i$ that represents the input features of the node $v_i$. If, for example, we iterate the  aggregation and update process of the node embedding for $k = 3$, the final learned representation of node $v_i$ is $\bm{z}_{i} = \bm{h}_i^{3}$, which includes the information about its 3-hop neighbourhood. 

To summarise, we have a step that creates a message for a vertex $v_i$ based on the knowledge about $N(i)$
\begin{equation*}
\bm{m}^{k}_{i} = \delta_{j \in N(i)}\big(\phi^{k}(\bm{h}^{k-1}_i, \bm{h}^{k-1}_j, \bm{l}_{i,j})\big),	
\end{equation*}
where $\delta$ defines a differentiable, permutation invariant function, e.g., sum or average, and $\phi$ is a differentiable function such as multilayer perceptron that creates messages between the vertex $i$ and the nodes in $N(i)$, incorporating the edge features $\bm{l}_{i,j}$. It is followed by the update step
\begin{equation*}
\bm{h}_i^k = \gamma^k(\bm{h}^{k-1}_i, \bm{m}^{k}_{i}),
\end{equation*}
where $\gamma$ specifies another differentiable function.

In the readout phase, we aggregate node features from the final iteration to obtain the entire graph representation $\bm{h}_G$
\begin{equation*}
\bm{h}_G = \zeta(\bm{h}^{K}_i, v_i \in V),
\end{equation*}
where function $\zeta$ should satisfy the same conditions as $\delta$, e.g. being invariant to graph isomorphism. This representation serves for the final task as, for instance, graph classification.

Depending on the types of graph convolutions (the creation and propagation of messages), we can categorise GCN into spectral-based and spatial-based models. The GCNs defined in the spectral domain (e.g. the Chebyshev Spectral GCN) are based on the graph Fourier transform, starting with the construction of the frequency filtering, whereas the spatial domain methods are specified directly on the graph, operating on groups of spatially close neighbours \cite{zhang2019comprehensive}. In the following section, we apply a  spatial-based GCN with the gaussian mixture model convolutional operator.

\section{Simulation study}
\label{sec:5}

To illustrate a possible combination of classical statistical tools together with machine learning algorithms for creating an enhanced network monitoring procedure, we consider an important and complex problem: the compliance of ambulance stations with the maximum response time.

\subsection{Motivation}
\label{sec:5.1}
There are examples of networks where failure could lead to irreparable harm. These networks can be defined as ``system-relevant'', where particular nodes represent ambulance or fire and rescue station. To guarantee proper functionality, these services are obliged to satisfy a strict policy regarding the time limit for arriving at the accident epicentre. For instance, in emergency medical cases, the ambulance must reach the patient without exceeding the legally prescribed response time. Its fundamental part is appointed to the travelling time and in some places is not allowed to exceed 12 minutes in 95\% of cases. The monitoring of the compliance with this rule can be performed by using the control chart for quantile function \cite{grimshaw1997control}. One of the potential choices is to create a test statistic based on the 0.8 and 0.95 quantiles so that possible deviation towards the maximum limit can be detected quicker.  However, in case of a signal, it remains unclear what led to its occurrence, unless we inspect the network state. One of the supportive methods in this task would be a GCN which can classify the network states in predefined categories, providing the first insight into a possible issue. This motivation guides the following application which is demonstrated on the simulated graph-structured data.

\subsection{Network definition}

Consider a simplified road network shown in Figure 1 whose topology is based on an existing city map. It can be graphically represented by $|V| = 18$ vertices and $|E| = 25$ edges. To differentiate between an ambulance node and a patient region, we introduce the vertex attribute ``Role''. We assume that the ambulance station can serve only one patient at once and that in total two fixed vertices define the ambulance stations. Also, which patient needs help from an ambulance is decided randomly. Thus, we include another vertex attribute ``Involvement in an accident'' that describes whether an ambulance station provides help (in this case, the value is set to 1) or is free (the value equals 0). Considering the patient nodes, as soon as the patient is involved in an accident, the value is set to 1. It remains 0 if no help is needed or obtained from the ambulance service. Both attributes are contained in $\bm{X}^V \in \mathbb{R}^2$ and can be found in Table \ref{Attributes}.

Regarding the edges, we model also two features $\bm{L}^E \in \mathbb{R}^2$ which reflect distinct types of roads. The first continuous attribute defines travelling time in minutes  $L^E_1$ (we select 1, 2, 3 and 5 minutes to be the expected values of passing respective roads) which is generated by applying lognormal distribution with different $\mu$ and $\sigma$. The selected values of $\mu$ and $\sigma$ parameters displayed in Table \ref{Attributes} reflect the target state of the network. The second attribute $L^E_2$ defines the level of construction works on the roads. Here, the in-control state is dominated by the attribute being equal to 0 or 1, that means no or minor roadworks are observed.  
\begin{figure}
	\begin{center}
		\includegraphics[width=1.0\textwidth, trim= 10 50 5 50,clip]{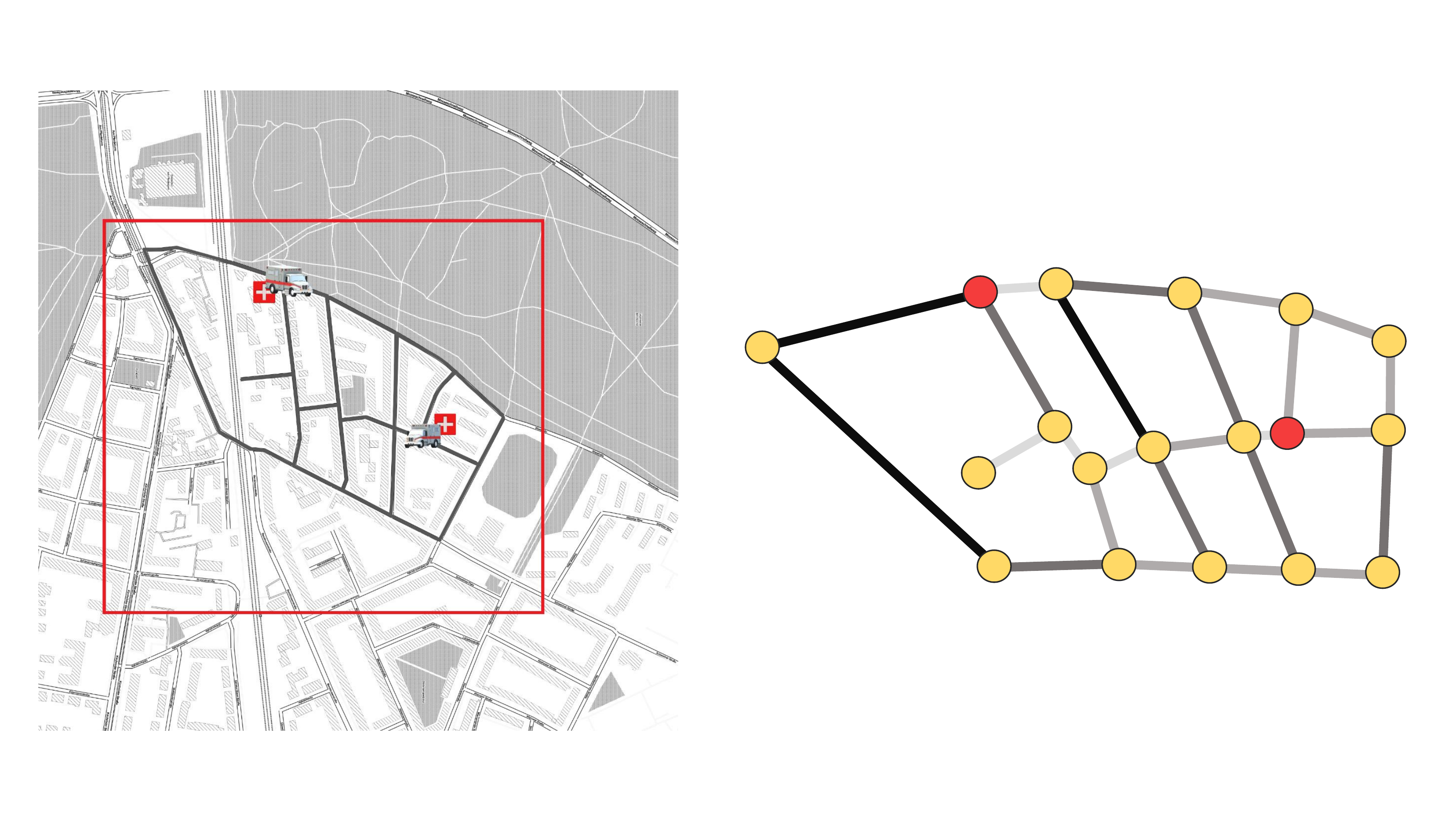}
		\caption{An example of a small road network which is derived from an existing city map with two arbitrary placed ambulance stations (left) and its undirected graph representation (right). Different colours of edges replicate the travelling time along the road, where a darker colour means longer travelling time. The red nodes define ambulance stations. }
		\label{fig:GraphRepres}
	\end{center}
\end{figure}
\begin{center}	
	\begin{table}[bp]
		\centering
		\small
		\begin{tabular}{|c|c|c|c|c|}
			
			\hline
			\multicolumn{1}{|c|}{Type} &
			\multicolumn{1}{c|}{Attribute} &
			\multicolumn{1}{c|}{Value}  \\
			\hline
			\makecell{Edge}&\makecell{Travelling time \\ (in minutes)}  & \makecell{$\EX(1)$ $\sim \mathscr{F}(0.1, 0.05^2)$ \\ $\EX(2)$ $\sim \mathscr{F}(0.7, 0.05^2)$ \\ $\EX(3)$ $\sim \mathscr{F}(1.1, 0.05^2)$\\$\EX(5)$ $\sim \mathscr{F}(1.6, 0.05^2)$ \\ with $\mathscr{F}(\cdot) = \mathrm{Lognormal}(\mu, \sigma^2)$} \\
			\cline{2-3}
			&\makecell{Level of road blocking \\ due to construction work} &\makecell{Free: 0 \\ Low: 1 \\ Middle: 2\\High: 3 } \\
			\hline
			\hline
			Node&Role &\makecell{ 0: Patient\\ 1: Ambulance}  \\
			\cline{2-3}
			&Involvement in an accident &\makecell{ 0: No involvement\\ 1: Help is provided, obtained or needed}  \\
			\hline
		\end{tabular}
		\caption[attr]{Edge and nodal attributes.}
		\label{Attributes}
	\end{table}
\end{center}
\subsection{Generation of the response time data}
For monitoring the travelling time from the ambulance station to the patients, we make some assumptions considering the simulation of the response time data. Daily, as soon as there is a patient call with the need for help, the travelling time of the ambulance which is closer to a patient is registered together with the current network situation. Sometimes, the number of accidents can be higher than one at the same time, so that the network situation is captured once in this case. However, if the network is in control, the maximum number of simultaneous accidents equals two, otherwise, there exists a personnel shortage. We assume that the ambulance follows the most efficient route, which is the shortest path in terms of travelling time between the ambulance station and the patient. For its calculation, we apply the Dijkstra's algorithm.

By the end of the day, the recorded response times are collected so that the 80 \% and 95 \% quantiles can be derived for defining the test statistic. If the test statistic exceeds the control limit, the collected network data are provided to the trained GCN that classifies the scenes into four different groups: a stable condition of the road network (label 0), an unstable condition due to the manpower shortage (label 1), an unstable condition due to the construction works (label 2) and an unstable condition due to the traffic jams (label 3). It is important to include the label 0 graphs which dominate in the definition of the in-control state for the identification of possible false alarms. To proceed with the application of the control chart itself, we first will explain how different label groups were designed.

\subsection{Road conditions}
\label{sec5.4}
To discuss the four condition classes, we should concentrate on the problem from the angle of what the neural network should learn in terms of main differences between the specified scenarios. Considering the reliable state, the neural network needs to distinguish between the problems on the roads which affect the travelling time so that it potentially lead to the out-of-control case and which not, as the patient was still reached on time. It means, despite the obstacles on the roads that can be modelled by increased values of edge attributes, each time the patient was reached by an ambulance in or under 12 minutes (simply because the route to the patient was not affected considerably),  the graph obtains the label ``0''.  

The class with the label ``1'' defines the problem of manpower shortage, meaning the reason of longer travelling times is due to the imbalance in the capacity of the ambulance service and the number of patients who needs help. In this case, we generate a higher number of patients, i.e., more nodes with ``Role = 0'' are involved into an accident ( $X^V_2 = 1$). As soon as both ambulance stations provide help and some further patients are not treated yet, the travelling time to these nodes is calculated as $\EX(L^E_1)$ $ \cdot 2 \cdot 2$, where $\EX(L^E_1)$ defines the expected value of edge feature ``Travelling time'' multiplied with the number of roads to pass on average and the need to travel first to the ambulance station and then to the patient. In this case, the network is out-of-control due to the considerably increased travelling time to the third and further patients.

Creating the unreliable situation on roads due to construction works (label 2), a particular group of roads which come from one or several distributions considering the travelling time is selected and higher values of the second edge attribute $L^E_2$ are assigned to these roads. Thus, the $\mu$ and $\sigma$ parameters are changed so that the higher attribute value corresponds to longer travelling time along the road.

For modelling the traffic jam (label 3), $L^E_2$ is set low (0 or 1) and some values describing travelling time of specific roads are generated from a different distribution that implies their increase. The examples from the groups with the labels 2 and 3 do not necessarily lead to an out-of-control state if the patients still reached under the critical time prescription.

After defining the network composition and specifying possible in- and out-of-control scenarios, we can collect the quantile observations for the calibration of the control chart and the graph representations of different classes, in order to train the neural network. 

\subsection{Calibration of the control chart for quantile function values}
\label{sec5.5}
To calculate daily 80\% and 95\% quantile values, we randomly simulate between 10 and 100 accidents which are repeatedly assigned to different patients. Next, the shortest paths between an available ambulance station and the selected patient are found and saved. Using the control chart presented in \cite{grimshaw1997control}, the test statistic is defined as follows  
\begin{equation*}
a_t = (\hat{\bm{Q}}_t - \bm{Q}_0)'\bm{\Sigma}^{-1}_0(\hat{\bm{Q}}_t - \bm{Q}_0),
\end{equation*}
where $\hat{\bm{Q}}_t = \big(\hat{Q}_{0.8,t}, \hat{Q}_{0.95,t}\big)'$ and the length of $\hat{\bm{Q}}_t$ is denoted by $c$. In Phase I, the expected value $\bm{Q}_0$ is estimated by the mean $\bar{\bm{Q}}$ and $\bm{\Sigma}_0$ by the sample covariance matrix $\bm{S}$ with 2500 in-control samples. 
For sufficiently large number of samples, $a_t$ follows the $\chi^2$ distriburion with $c$ degrees of freedom, if the sample at time point $t$ corresponds to the specified in-control state. Hence, the control limit can be defined by $\chi^2_{\alpha}(c)$, selecting $\alpha$ with respect to the in-control average run length (ARL) using $\alpha = 1/ARL$. Here, we choose ARL = 1000, therefore, $\chi^2_{0.001}(2) = 13.82$.

\subsection{Construction and training of the graph convolutional network}

In this simulation study, we are interested in the classification of collected graphs which belong to a change point. The goal is to assign a given graph to one of the predefined categories by learning the feature representation from provided training data which contain class labels. Consequently, we have to define the GCN architecture so that it can solve the specified task. Also, our graph convolutional operator should be capable to integrate the node, as well as edge, attributes into the message passing process because they encompass valuable information about the network's condition.

\begin{figure}
	\begin{center}
		\includegraphics[width=1.0\textwidth, trim= 0 0 0 0,clip]{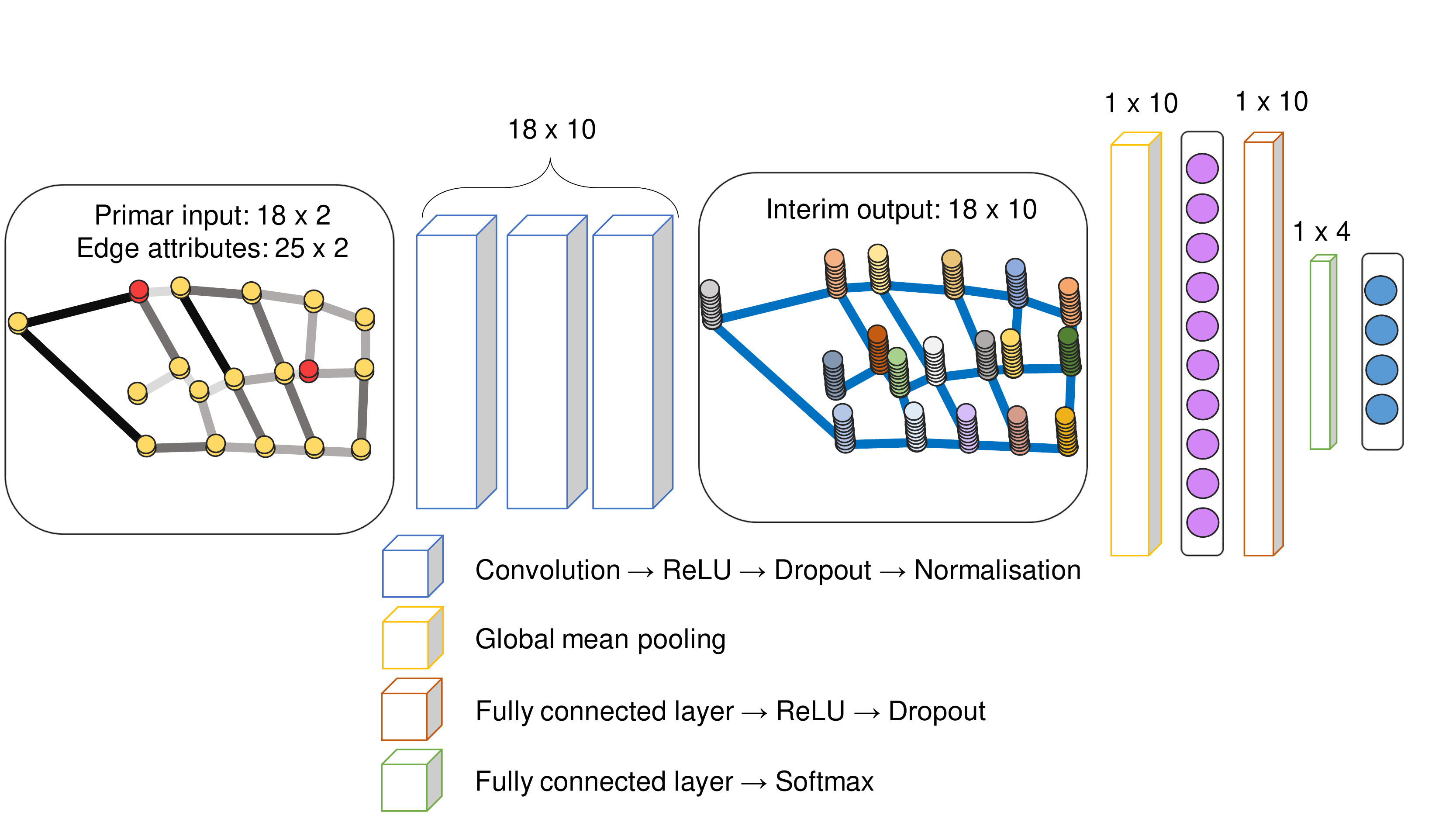}
		\caption{The schematic architecture of the applied GCN. Each block represents a single layer where the first stage (blue blocks) contains graph convolutional layers with layer normalisations for learning the feature representation and the second stage (yellow, orange and green blocks) consists of dense layers for classification.}
		\label{fig:GCNArchitecture}
	\end{center}
\end{figure}

Figure \ref{fig:GCNArchitecture} presents the architecture of the applied GCN. The first three graph convolutional layers, each encoding the input in a feature vector of size $18\times10$, perform three propagation steps and effectively convolve the 3rd-order neighbourhood of every node. 
We chose the gaussian mixture model convolutional operator \cite{monti2017gaussiangcn} which is implemented in the programming framework provided by \cite{fey2019pytorch}. Each convolution step is followed by the rectified linear unit (ReLU) activation function. Afterwards, the dropout operation is applied which randomly sets the processed input units to 0 with a specified frequency $\xi$ (in our case, $\xi$ = 25\%) during the training time, preventing the model from overfitting, i.e., learning from the training dataset without its generalisation. Before the next convolution begins, we normalise the inputs across the features that is known as layer normalisation (cf. \cite{ba2016layernorm}).

After the message passing phase, a readout layer that is defined by a global mean pooling operation transforms the latent vertex representations to a graph representation as a fixed-size vector. Here, the interim output is averaged across each hidden node dimension so that the graph-level output size is $1 \times 10$. Next, we attach two fully connected layers to increase the ability to learn a complex function. The second layer predicts the final class probability distribution of size $1\times4$ followed by the softmax activation function.
We cannot apply the ReLU activation function as it provides continuous output in range $[0; \infty]$. In the final stage, we need the output to be in the finite range $[0; 1]$ for interpreting its results as probabilities, with the highest value corresponding to the predicted class. 

After defining the architecture of the GCN, we can start with training or fitting the neural network. This procedure involves usage of a training dataset to update the model parameters (weights and biases) so that we obtain a reliable mapping between input (graph) and output (class label). For the training dataset, we generate 2500 graphs. It is important to avoid class imbalance during the training process, therefore, each label is represented by the same number of examples. Another vital part of the training process is the loss function. It calculates the difference between the computed output from the input data (this process is known as ``forward pass'') and the value provided as ground truth. Here, we choose the negative log-likelihood loss which is appropriate for a multiclass classification problem. It defines our objective function which we minimise by updating the model parameters.

The results which are provided by the loss function are applied in the optimisation step of our parameters that is based on gradient computation (known as ``backpropogation'' or ``backward pass''). The negative log-likelihood is minimised using the Adaptive Moment Estimation (Adam) function \cite{ba2016adam} with a learning rate of $10^{-3}$.

\begin{figure}[t]
	\centering
	\begin{minipage}{.5\textwidth}
		\centering
		\includegraphics[width=1.0\textwidth, trim= 40 80 5 120,clip]{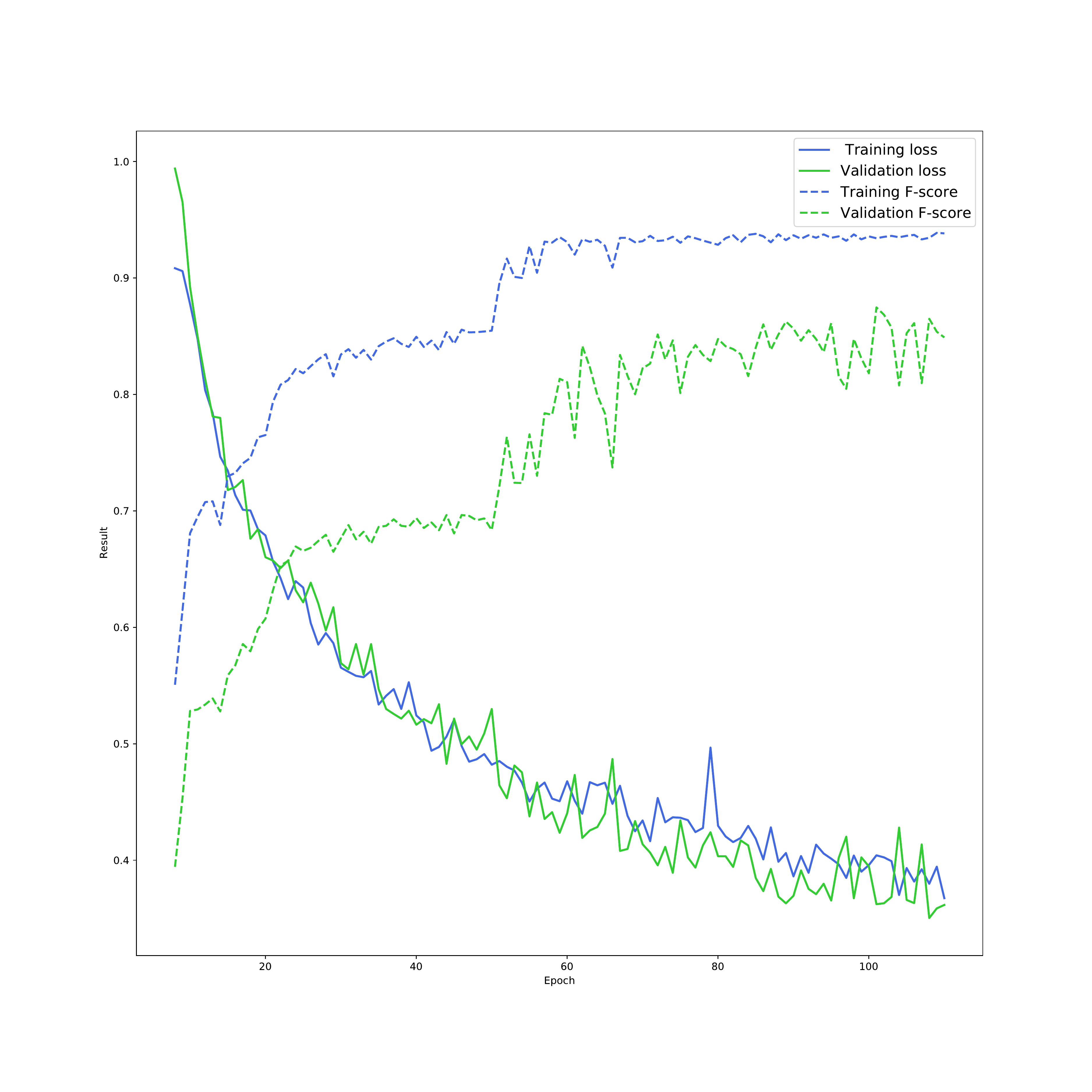}
		\label{fig:test1}
	\end{minipage}%
	\begin{minipage}{.5\textwidth}
		\centering
		\includegraphics[width=1.2\textwidth, trim= 30 20 5 50,clip]{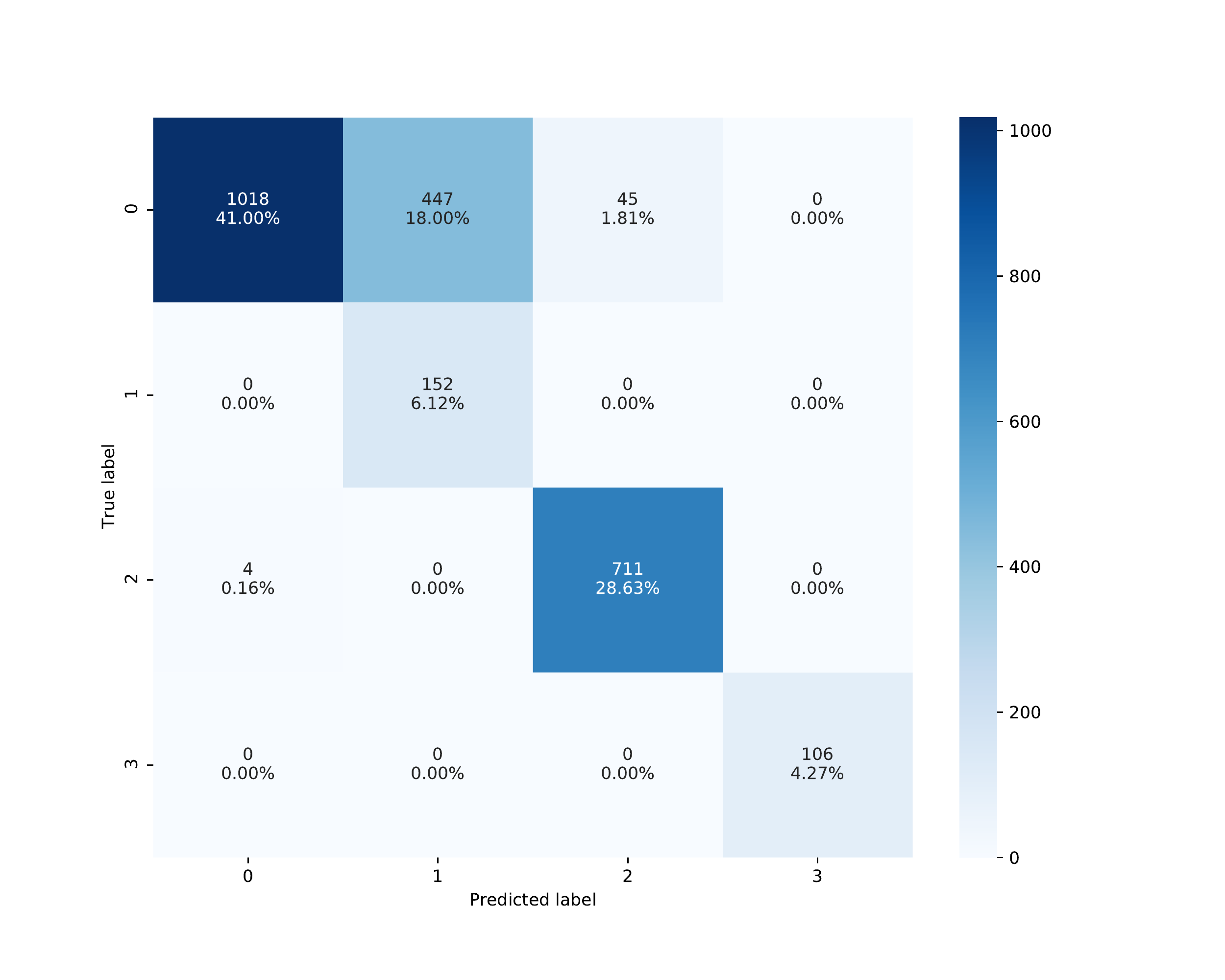}
		
		\label{fig:ConfMatrix2}
	\end{minipage}
	\caption{The training progress shown on the training (blue curves) and validation sets (green curves) (left). The confusion matrix presenting the performance of the trained GCN in Phase II (right). The numbers on the diagonal represent the proportion of correctly classified examples (compared to the size of the complete test dataset) and the off-diagonal entries correspond to the proportions of the misclassified graphs.}
	\label{fig:PerformanceMeasures}
\end{figure}

The execution of the backward and forward pass together defines one iteration. During one iteration, we usually pass a subset of the dataset known as ``mini-batch''. In case we decide to pass all data at once, it is called ``batch''. Here, we train the neural network using the mini-batch with size 16, i.e., every iteration 16 graphs are processed together. As soon as the entire dataset was passed, one epoch is completed.
As a performance metric which supports the selection of the best model, we compute weighted F-score after each epoch. Figure \ref{fig:PerformanceMeasures} illustrates the training and validation history of the applied GCN. 
To test how well the network generalises to unseen data, we apply the holdout validation method. The validation set, which contains more complex samples, i.e., completely new examples which belong to the classes but are not included in the training dataset, was designed with a size of 800 graphs.
In order not to overtrain the network, we use early stopping with respect to the F-score improvement, which terminates the training process if the value has not increased within 10 epochs. We find the optimal model to be at epoch 102 with 93\% and 87\% the weighted F-Score of the training and the validation dataset, respectively. However, to see whether the model functions certainly correctly, we need to test it on a new dataset coming from the monitoring procedure.  

%

\subsection{Phase II analysis}
\begin{figure}
	\begin{center}
		\includegraphics[width=0.85\textwidth, trim= 60 30 30 60,clip]{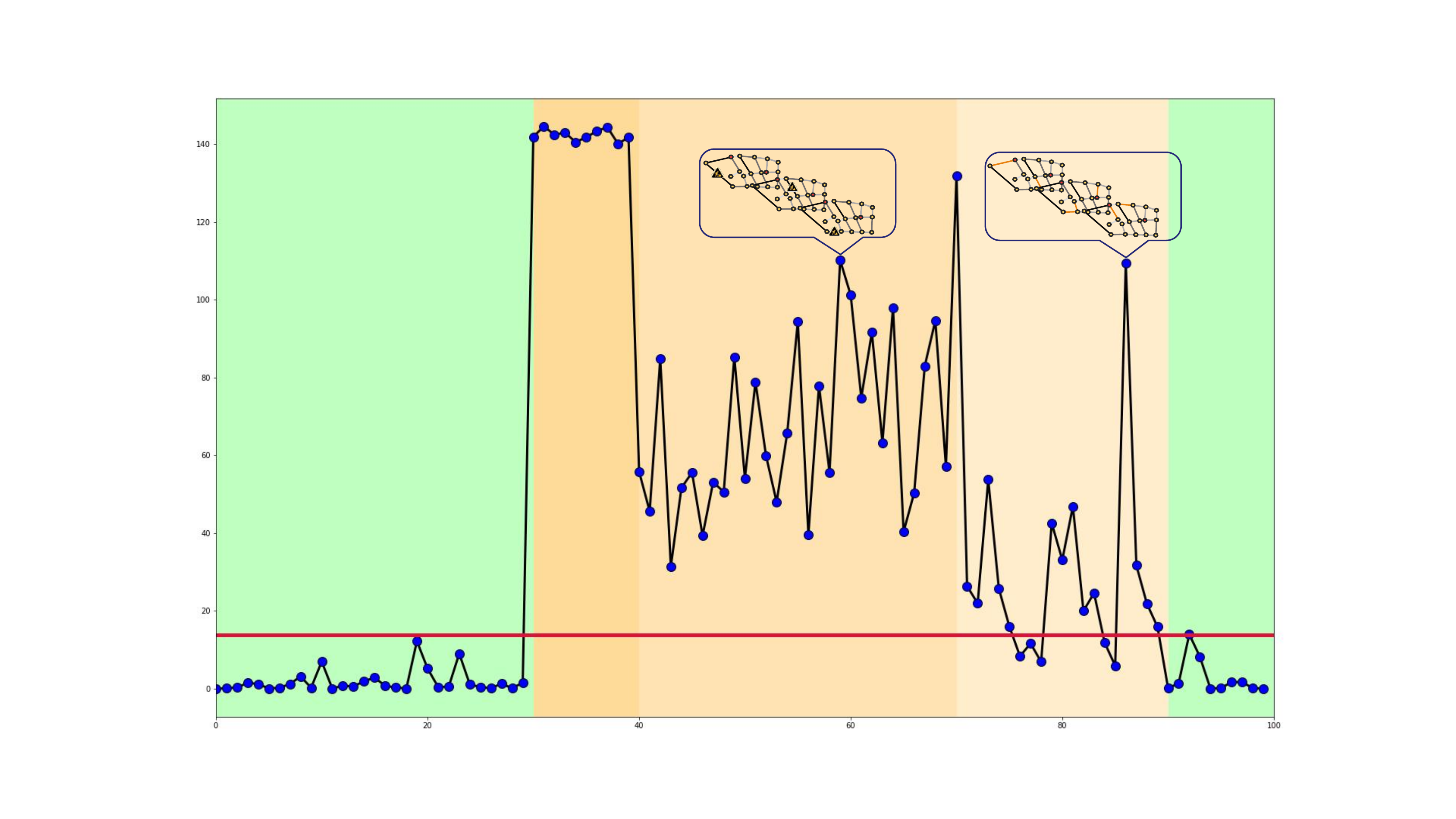}
		\caption{The control chart for quantile function values. The horizontal red line corresponds to the
			control limit. The green areas are designed by the label 0 cases, the dark orange by the label 1, followed by the label 2 and 3. The incorporation of the graphs with different properties such as construction works
			(triangular symbols) or traffic jams (orange coloured edges) defines the availability of additional
			information to understand the reason of the detected change point.}
		\label{fig:ControlChart}
	\end{center}
\end{figure}
Here, we combine the implementation of Phase II with testing the trained GCN. We define the length of the monitoring period to be 100 days, where the network in the first 30 and the last 10 days is considered to be in control. The out-of-control period is designed in the remaining days, where the process is exposed to the personnel shortage (10 days), excessive construction works (30 days) and increase of traffic jams (20 days). After simulating the cases and calculating the quantiles, we obtained the control chart presented in Figure \ref{fig:ControlChart}. In terms of false signals, there is one in the last 10 days which slightly exceeds the control limit. The possible reason may be the high variance in the in-control data. We can also notice that not all the test statistics show the out-of-control state in the period when the network was exposed to an increased number of traffic jams, however, they do not define missing signals. As it was mentioned in Section \ref{sec5.4}, if the ambulance services were still able to reach the patients within the allowed time, then no out-of-control state is given.

Normally, we would apply the neural network only in the out-of-control state, however, here the primary aim is to evaluate the performance of the trained GCN to classify provided graph observations in general. Hence, we create a test dataset using the data from Phase II, i.e., from the 100 generated days which include both in-control and out-of-control periods and examples from each of the four classes. As we can see in Figure \ref{fig:PerformanceMeasures}, the GCN can almost flawlessly identify the classes 0, 2 and 3. However, the class 1 seems to be not well learned, possibly due to lacking clarity in its representation. Overall, the model achieves the weighted F-score of 83\% being an encouraging result.

\section{Conclusion and discussion}
\label{sec:6}

In many applications, treating the underlying data as a graph can achieve greater efficiency. However, data representation in the form of graphs is still novel for both machine learning and statistics. Hence, it is particularly important to use synergy effects between two different statistical learning frameworks to develop efficient and modern analytical approaches. In this paper, we uncover the possibility to bring together statistical process control and deep learning algorithms to monitor graph-structured data. 

Learnable models which operate on graphs are only a stepping stone on the path toward a significant expansion in understanding the environment. Besides the topic of how to unify both frameworks, from the statistical perspective, there are many other open questions in this area. How to represent the graph data and convolve the information, which approach to use in which case and how to measure its performance: these and many other challenges are yet to be conquered.

It is a natural question, to consider whether one could expand the use of application of algorithms such as GCNs to encompass the whole monitoring procedure, omitting control charts altogether.  Although an appealing idea, the complexity of the model required for real-world data, combined with the amount of training time necessary, would severely limit the applicability of such an approach. Applying a hybrid method allows us to take advantage of the efficiency of classical methods, while using modern machine learning in order to specify more subtle network characteristics, which normally require human-lead scrutiny to determine.

We believe that a better way to understand the relationship between both frameworks is that machine learning is the logical next step in response to the growing volume of data. Thus, it is beneficial to see the successes in artificial intelligence application not as an attempt to replace the traditional statistical methods but as a direction towards their enhancement.

\end{document}